\documentclass[letterpaper,10pt,conference]{ieeeconf}
\IEEEoverridecommandlockouts

\usepackage[space, compress, sort]{cite}
\usepackage{amsmath,amssymb,amsfonts}
\usepackage{graphicx}
\usepackage{textcomp}
\usepackage{xcolor}
\usepackage{booktabs}
\usepackage{multirow}
\usepackage{url}
\usepackage{tikz}
\usepackage{todonotes}
\usepackage{caption}
\usepackage{changes}
\usepackage[breaklinks=true,bookmarks=true,colorlinks]{hyperref}
\usetikzlibrary{arrows.meta,positioning,fit,backgrounds}

\newif\ifanon

\title{Learning to Drive on Mars: Visual Multimodal Traversability Estimation for Off-World Navigation}
\begin{document}

\ifanon
  \author{Anonymous Authors}
\else
  \author{Darren Chiu$^{1,3}$, Cole Wilson$^{2,3}$, Andrei Tumbar$^{3}$, Gaurav S. Sukhatme$^{1}$, Steven Myint$^{3}$%
    \thanks{Corresponding author: chiudarr@usc.edu}%
    \thanks{$^{1}$University of Southern California, Los Angeles, CA 90089, USA.}%
    \thanks{$^{2}$Washington State University, Pullman, WA 99164, USA.}%
    \thanks{$^{3}$Jet Propulsion Laboratory, California Institute of Technology, Pasadena, CA 91109, USA.}%
    }
\fi

\maketitle
\thispagestyle{empty}
\pagestyle{empty}
\begin{abstract}
    Autonomous navigation on Mars requires vehicles to distinguish between traversable terrains across diverse and visually challenging environments. However, progress in learning-based navigation for off-world environments has been limited by the lack of large-scale datasets. Since landing in Jezero Crater, the Mars 2020 Perseverance rover has traversed terrain ranging from sandy dunes, rocky patches, and flat bedrocks. As a result, this paper presents a dataset spanning 500 sols and 45km of trajectories driven by both human operators and the onboard planner, ENav. Our dataset contains grayscale stereo image pairs, poses, accelerometer readings, rocker-bogie angles, and estimates of tilt and wheel slip. Building on this dataset, we introduce an uncertainty-aware traversability-estimation framework that learns terrain representations from multimodal driving experience. We compare our proposed method against existing approaches on the Mars 2020 dataset and show that our method achieves an AUROC of 0.874 and an F1 score of 0.758, outperforming the strongest baseline by 0.058 and 0.156, respectively, while also achieving the highest average precision and recall. Finally, we show that the visual representations can be integrated into path planners, such as ENav, on a physical rover test bed. Videos, code, and the M2020 dataset will be available \href{https://darren-chiu.github.io/learning-to-drive-on-mars/}{here}. 
\end{abstract}

\section{Introduction}
Autonomous navigation is a fundamental capability for robotic exploration of planetary surfaces. Unlike terrestrial robots, planetary rovers operate under severe communication constraints, where long command cycles make continuous human supervision impractical and expensive. Consequently, planetary rovers increasingly rely on onboard autonomy to perceive terrain, estimate their state, and plan safe trajectories. The Mars 2020 Perseverance rover represents a significant step toward this vision: its autonomous navigation system, ENav, constructs 2.5D maps from stereo imagery and provides geometrically safe trajectories. Throughout its operation, ENav was used to autonomously drive the majority of Perseverance's journey \cite{verma_ops}.
Despite these advances, terrain assessment for planetary rovers remains largely grounded in geometric reasoning. Stereo cameras provide a powerful source of spatial information, allowing the rover to detect obstacles, estimate terrain geometry, and reason about vehicle clearance. However, geometric planning alone does not fully capture the interaction between a vehicle and its environment, such as the potential for slippage. Terrain that appears geometrically traversable may produce substantial wheel slip, unexpected vehicle motion, or unfavorable interactions with the rover's mobility system. These effects are particularly important for planetary vehicles, where recovery from a mobility failure is impossible.

Deep learning offers an opportunity to learn these terrain-dependent effects directly from driving experience. A growing body of work has explored supervised and self-supervised traversability estimation for ground robots, using proprioceptive signals such as acceleration to generate supervision for visual terrain models \cite{Castro_2023_HowDoesItFeel}. More recently, pretrained visual representations and visual foundation models have demonstrated promising transfer across visually diverse environments, reducing the amount of task-specific data required to learn representations of traversability \cite{Triest_2025_Velociraptor}. These developments suggest that rover driving telemetry and proprioception could provide useful labels about terrain suitability in addition to conventional geometric hazard labels.
\begin{figure}[!tbp]
    \centering
    \includegraphics[width=0.99\linewidth]{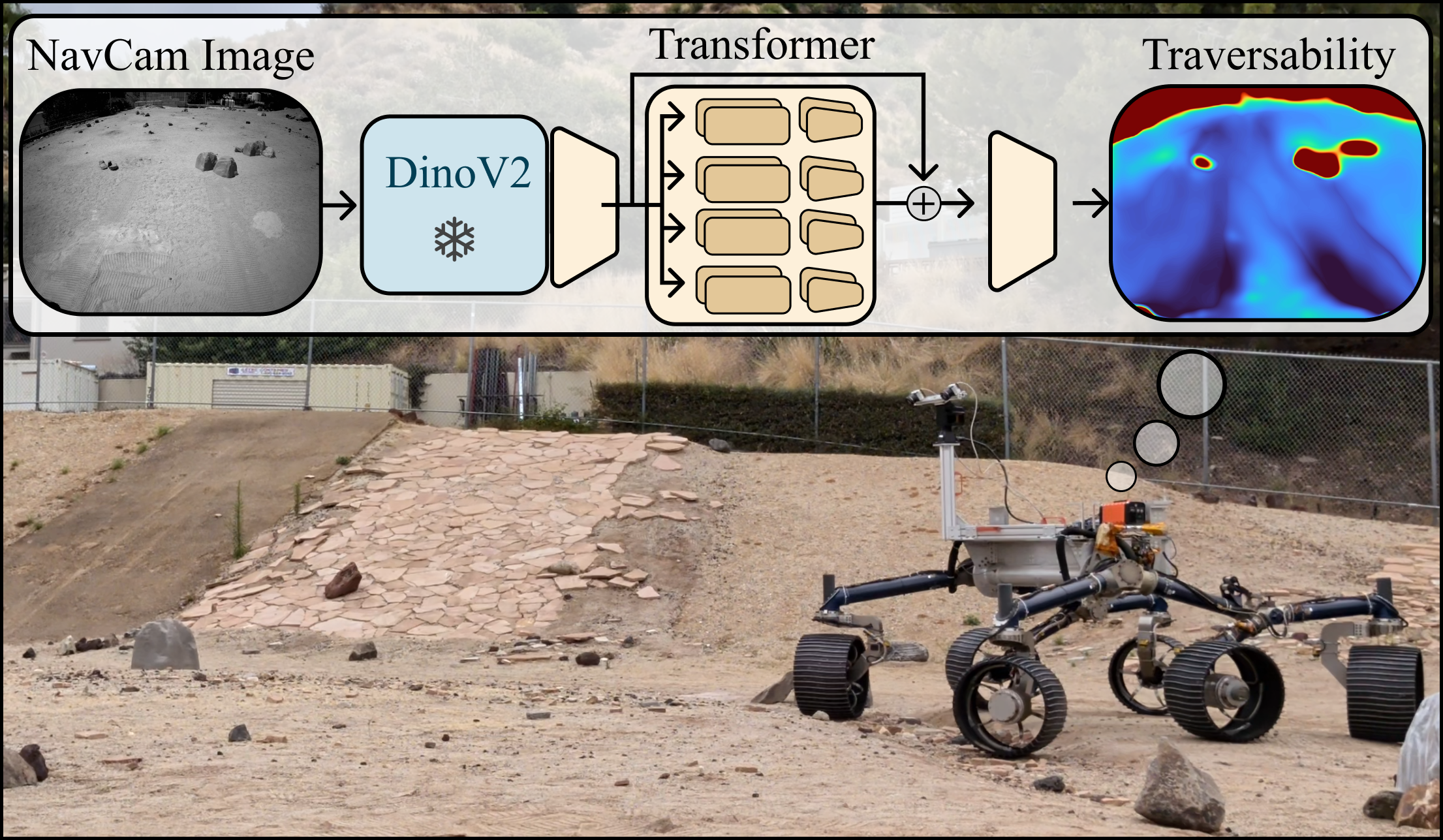}
    \caption{The Perseverance rover's Earth analog deployed in a Martian-like rock field. Our traversability estimation model is trained entirely on Martian driving data and deployed directly with no finetuning on images here on Earth.}
    \label{fig:hero_fig}
    \vspace{-1.5em}
\end{figure}
However, planetary robotics lacks a large-scale dataset that directly connects driving experience with the multimodal observations required for learning traversability. Existing datasets surrounding Mars have primarily focused on visual semantic or terrain classification. For example, AI4MARS provides hundreds of thousands of semantic segmentation labels over imagery collected by the Spirit, Opportunity, and Curiosity rovers \cite{ai4mars}. While these annotations provide valuable information about visual terrain appearance, they do not directly encode how a six-wheeled rover physically interacted with the terrain. 

In this paper, we introduce a large-scale dataset collected from approximately 500 sols of Perseverance surface mobility operations. The dataset provides grayscale navigation camera (NavCam) stereo observations with rover poses, inertial measurements, rocker-bogie configurations, and estimated slip and tilt values, providing a multimodal record of both what the rover observed and how it responded to the terrain. Importantly, the trajectories represented in the dataset are not artificially generated: they arise from a combination of human Rover Planner decisions that are executed in conjunction with the onboard ENav software. The resulting data therefore captures the distribution of terrain, vehicle states, and driving conditions encountered during actual planetary exploration.

Built upon this dataset, we investigate whether off-world driving experience can be used to learn a visual representation of terrain traversability. We leverage visual foundation models to extract transferable features and train a traversability estimator using supervision derived from the rover's proprioceptive data. The learned representation is subsequently projected into a traversability map that is integrated into the existing stereo-based planner, ENav. Rather than replacing the geometric autonomy stack, our approach instead augments it with a learned assessment of terrain.

We validate the resulting system on an Earth equivalent rover platform with weight and mobility characteristics matching Perseverance on Mars. This evaluation provides a physical test of whether information learned from Martian driving data can improve navigation in closed-loop scenarios. Our contributions are threefold:
\begin{enumerate}
\item We introduce a 500-sol ($45km$) multimodal driving dataset collected from the Mars 2020 (M2020) Perseverance rover, containing synchronized stereo imagery, poses, inertial measurements, rocker-bogie angles, along with tilt and slip estimates.
\item Built on this data, we present a contrastive learning framework that leverages pretrained visual foundation models and rover driving experience to estimate terrain traversability from visual observations.
\item We demonstrate closed-loop integration of the learned traversability representation with the ENav \cite{toupet_2026_enav} planner on an Earth equivalent rover.
\end{enumerate}
Together, the dataset and proposed learning framework establish a connection between operational planetary rover data and modern deep learning-based navigation, enabling future research in autonomous planetary exploration and navigation.
\section{Related Work}

\subsection{Planetary Rover Datasets and Perception}
The development of learning-based methods for planetary exploration has been constrained by the limited availability of real planetary data. AI4MARS \cite{ai4mars} and SPOC \cite{Rothrock2016SPOCDL} introduced a large-scale dataset of approximately 35K images and 326K semantic segmentation annotations collected from the Spirit, Opportunity, and Curiosity rovers. This dataset has enabled research on terrain classification and terrain-aware perception, but its labels primarily describe visual terrain categories rather than the physical suitability of terrain for rover traversal. 
These datasets demonstrate the value of rover imagery for learning-based perception, but do not provide a long-duration multimodal record linking visual observations to rover state and terrain interaction. Our dataset complements these efforts by associating stereo imagery with pose, inertial measurements, rocker-bogie configuration, and slip estimates collected during actual Perseverance operations.

\subsection{Autonomous Navigation for Planetary Rovers}
Planetary rover navigation has traditionally relied on geometric planning or human-in-the-loop operations. Stereo imagery can be used to construct maps from which obstacles, slopes, and other geometric hazards are identified. Perseverance's autonomous navigation system, ENav, leverages stereo-derived depth maps allowing the rover to autonomously select and execute safe paths over unseen terrain \cite{toupet_2026_enav}. Whereas feature matching can be done to reduce the localization uncertainty \cite{verma_localization}. Geometric representations provide important safety guarantees and are particularly well suited to where training data is scarce. However, geometric properties alone do not necessarily capture the interaction between the rover and terrain. Factors such as wheel slip, soil compliance, and vehicle configuration can affect traversal quality even when terrain is geometrically feasible. Our work therefore considers learned traversability as a complementary signal that can be integrated with an existing geometric navigation pipeline rather than replacing it.
\subsection{Learning-Based Terrain Traversability}
A growing body of work has investigated learning traversability directly from a robot's experience. Traditional traversability estimation often computes costs from manually designed geometric features such as slope, roughness, and obstacle height. Learning-based methods instead seek to infer terrain suitability from observations collected from driving datasets \cite{anelia_visualslip, yumi_ir_classification, Skonieczny_jfr, ono_riskaware, Patel_roadrunner, meng_terrainnet, Seo_off-road}, allowing the robot's physical interaction with the environment to provide supervision. Castro et al. \cite{Castro_2023_HowDoesItFeel} proposed self-supervised cost map learning in which proprioceptive terrain-interaction signals are used to supervise visual traversability prediction. Jeon et al. \cite{Jeon_2024_FollowFootprints} similarly exploit previously traversed regions to generate self-supervision for visual and geometric traversability estimation. Cunningham propose to use classification with geometric slope labels to predict slip using gaussian processes \cite{Cunningham_slip}.

These approaches are particularly relevant to planetary robotics because manual labeling of terrain according to rover mobility is impractical. The Perseverance driving logs provide a natural source of such supervision: visual observations can be associated with the rover's observed motion, inertial response, vehicle configuration, and perceived slip. This enables traversability to be defined in terms of the behavior response of the actual rover rather than a generic semantic terrain category. Our work extends experience-based traversability learning to real planetary driving data collected over hundreds of sols. 
\begin{figure*}[t!]
    \centering
    \includegraphics[width=1\linewidth]{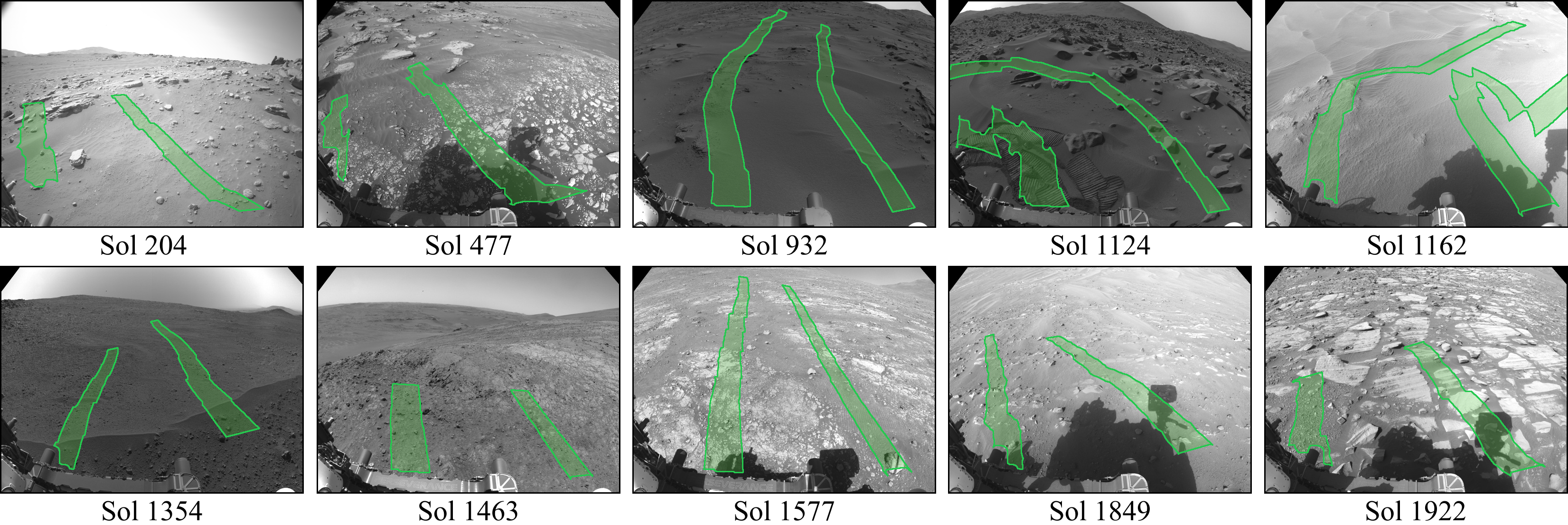}
    \caption{Visualizations of the dataset and wheel track projections from various Sols. The rover on Mars encountered various terrain types such as rocky hills, flat bedrock, sand ripples, and loose gravel. Our dataset contains $45km$ of total driving data and a cumulative $43,000$ total images.}
    \label{fig:dataset}
    \vspace{-1.5em}
\end{figure*}
\subsection{Visual Foundation Models for Navigation}
Recent advances in visual foundation models provide another opportunity to improve terrain understanding when task-specific labeled data are limited. Large pretrained visual models encode representations learned from diverse image distributions and have demonstrated strong transfer to downstream robotic perception tasks. In off-road navigation, recent approaches have investigated using these representations to infer traversability without requiring dense human annotations. Wild Visual Navigation \cite{mattamala25wild} uses pretrained visual features together with online self-supervision to learn traversability in previously unseen outdoor environments. Velociraptor \cite{Triest_2025_Velociraptor} further combines visual foundation models with geometric information to produce risk-aware traversability and cost representations for off-road navigation.

Our work builds on this emerging direction while addressing the visual domain gap between \textit{off-world} environments. 
Rather than relying solely on pretrained representations, we use Perseverance's driving experience to adapt visual features to the rover-specific notion of traversability. 
This combination of foundation-model representations with long-duration planetary driving data provides a mechanism for transferring the broad visual knowledge of Earth based pretrained models while grounding the resulting predictions in ground truth rover-terrain interactions.
\begin{figure*}[t!]
    \centering
    \includegraphics[width=0.95\linewidth]{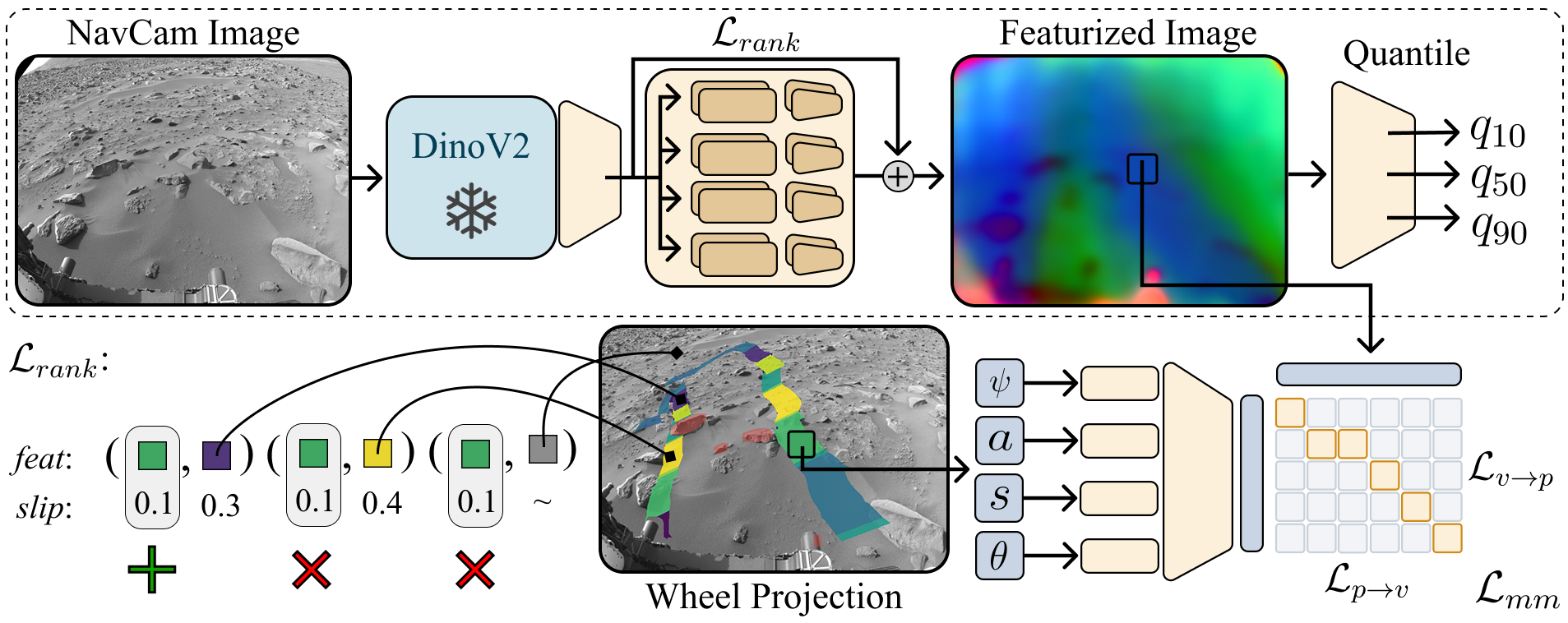}
    \caption{Architecture of the traversability model. The dotted region denotes the deployment path. The transformer component is trained through two core loss functions: a contrastive feature ranking noted $\mathcal{L}_{rank}$ and a contrastive multimodal alignment, $\mathcal{L}_{mm}$. $\mathcal{L}_{rank}$ is meant to rank images features based on how well the rover drove over it (measured by slip) and contrasts it from features that were not driven on ($\sim$), such as large boulders. The $\mathcal{L}_{mm}$ then contributes other proprioceptive signals such vehicle tilt, rocker-bogie angles, and acceleration. The final MLP is trained using a pinball loss to generate quantiles of slip prediction. Our final traversability score is generated by taking the $traversability = \max(q_{90} - q_{10},\, q_{50})$, where $q_{50}$ represents the median slip prediction and $q_{90} - q_{10}$ is the prediction uncertainty (predicted slip distribution width).}
    \label{fig:method}
    \vspace{-1.5em}
\end{figure*}
\section{The Mars 2020 Perseverance Dataset}
The dataset (examples shown in Figure \ref{fig:dataset}) comprises approximately 500 sols (Martian equivalent of a day on Earth) of driving sequences, for a total of $45 \ km$ of traversal and $43,000$ images. Each sol contains a sequence of grayscale image pairs, acceleration readings, rocker-bogie angles, tilt, and slip estimates. The grayscale image pairs are generated once every $1m$ of driving whereas pose and proprioception is collected at approximately $8 \ hz$. Figure \ref{fig:dataset} shows 10 geographically distinct images from the dataset where the green areas denote the projected wheel tracks (approximated through the two front wheels). The pose data is captured in the z-down coordinate frame in meters ($m$). The rocker-bogie ($\theta$) angles represent the 4 independent suspension angles in radians; these values are responsible for measuring the suspension's compliance over large rocks. The tilt ($\psi$) and slip ($S$) estimates are calculated and saved onboard. 
\begin{align}
    slip \equiv S = \frac{\| p_{wo} - p_{vo} \|}{dist_{wo}} \\
    tilt \equiv \psi =  cos^{-1}(cos(\phi) \cdot cos(\theta))
\end{align}
where $\phi$ denotes the current roll and $\theta$ the pitch. The $p_{wo}$ is the positional estimate derived from wheel odometry while the $p_{vo}$ is from visual odometry. As such, the $slip \in [0,1]$ estimate is a value that measures how much the wheels intended to move over what was \textit{visually} measured i.e. the unrealized commanded displacement. We find that this measurement is a good proxy for learning traversability.   
Because the martian environment is static, image acquisition is done only once every time the onboard navigation software plans a path. As a result, our training framework should take advantage of the sparse imagery and in comparison dense pose and proprioception.   
\section{Traversability Estimation Framework}
Our method learns a dense visual traversability estimator from data the rover previously collected on Mars. We accomplish this by formulating the problem as a feature ranking problem being solved in two parts. First, an ordered latent space is learned through a 5M parameter transformer. The transformer operates on the features obtained from a DINOv2 encoder \cite{oquab2023dinov2} and is trained with a modified Rank-N Contrast loss \cite{zha2023rank} and a cross-modal alignment loss inspired from the CLIP model \cite{clip}. Then, we fit an MLP on the transformer's latent space to produce per-pixel traversability metrics. Our dataset is split into three stages (accounting for geographic diversity within each set), a main stage for training the transformer, a small subset for fitting the quantile head, and a hold out set for experiments and evaluations (Section \ref{sec:results}).
\subsection{Model Architecture}
Our deployed model can be understood in three components: i.) a pretrained frozen DINOv2 (\textit{ViT-L/14}) encoder \cite{oquab2023dinov2} ii.) a learned 5M parameter transformer decoder that generates a slip organized space iii.) and a two-layer MLP head responsible for quantile prediction.  
\subsection{Contrastive Feature Ranking}
It is a rare occurrence that the Perseverance rover encountered large slip values. Therefore, it is difficult to naively learn regression models without more diverse and likely unsafe Martian traversal data. We find that our dataset lends itself cleanly to a ranking task, where we want to find which visual features slipped \textit{more} than others. Let $\{f_i\}_{i=1}^P$ denote features at wheel tracks for a sampled grayscale image with slip estimates $\{s_i\}$. $\{n_m\}$ will denote negative image features, i.e. rocks and sand ripples that the rover planners or onboard geometric planner \cite{toupet_2026_enav} intended to avoid. For anchor $i$ and positive $j$ the loss is described as:
\begin{align}
\mathcal{L}_{\text{rank}}
&= a \sum_i \sum_{j \neq i} -\log \frac{e^{\ell_{ij}}} {\sum_{k \in \mathcal{S}_{ij}} e^{\ell_{ik}} + \sum_m e^{\ell_{in_m}}},
\label{eq:rank}
\\
\mathcal{S}_{ij}
&= \{k \neq i : |s_i-s_k| \ge |s_i-s_j|\},
\\
a
&= \frac{1}{P(P-1)}.
\end{align}
Where  $\ell_{ij} = f_i \cdot f_j / \tau$ and $\ell_{in_m} = f_i \cdot n_m / \tau$ denote the similarity scores between the positive and negative pairs.
Following V-STRONG \cite{vstrong}, we use SAM prompts within the wheel tracks to expand the sparse visual signals.  
\begin{align}
    \mathcal{L}_{\text{mask}} = \frac{1}{P'} \sum_{i=1}^{P'} \Bigg[
\log \sum_m e^{f'_i \cdot n'_m / \tau}
\;-\; \frac{1}{P'-1} \sum_{j \neq i} \frac{f'_i \cdot f'_j}{\tau} \Bigg],
\end{align}
where  $\{f'_i\}_{i=1}^{P'}$ denotes features that were expanded from segment anything prompts \cite{kirillov2023segany}. Our total ranking loss is then denoted as:
$$
\mathcal{L}_{pix} = (1 - \omega_{\text{mask}})\,
\mathcal{L}_{\text{rank}} + \omega_{\text{mask}}\, \mathcal{L}_{\text{mask}},
$$
For our experiments, we use $\omega_{\text{mask}} = 0.05$.
\subsection{Multimodal Alignment}
We find that applying the ranking loss described in equation \ref{eq:rank} is not suitable for a general traversability estimation task as the objective only considers slip differences. We find that the visual features generated from our transformer block are improved by grounding the features using a modified approach from the CLIP models \cite{clip}. This stems from the idea that many proprioceptive readings should contribute to the vehicle slip, such as tilt and rocker-bogie angles. For each batch we sample $S$ proprioception rich segments of the wheel track (each $1m$) and pair them with $N$ visual features, where $N>>S$. 

Within our dataset, we encode the power spectral density of the tilt, slip, rocker-bogie angles, and acceleration values into a $p_{embed} \in \mathbb{R}^{64}$ vector through a 2 layer MLP to generate $z_s$. Each $1m$ segment contains multiple visual features, but only one proprioceptive embedding. Similar to CLIP we begin by forming a logit matrix of visual features and the proprioceptive embeddings. 
\begin{figure*}[t!]
    \centering
    \includegraphics[width=0.95\linewidth]{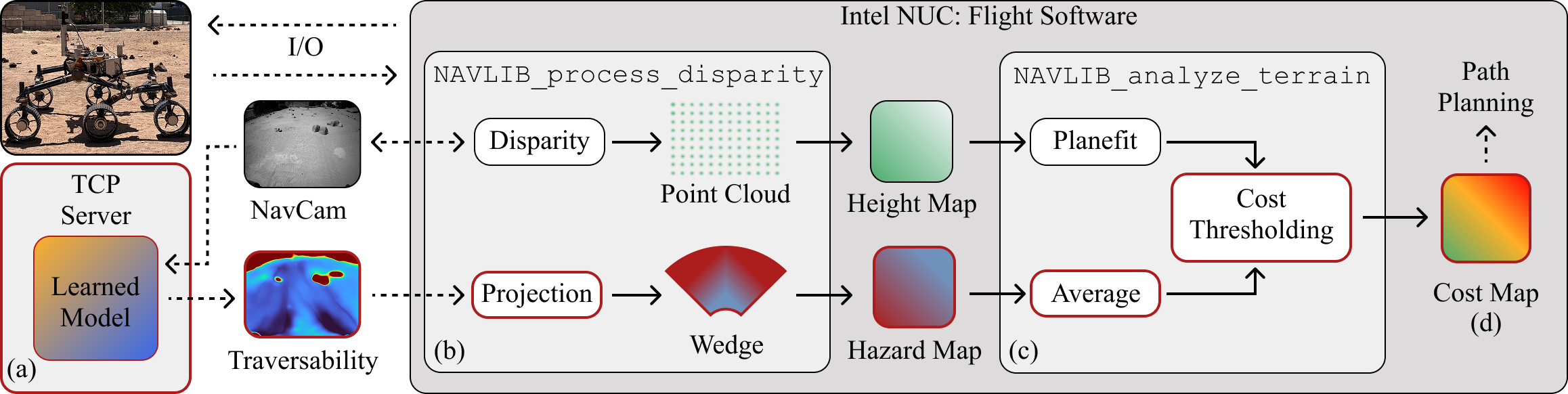}
    \caption{Integration into the Mars 2020 ENav \cite{toupet_2026_enav} FSW on our rover testbed; red borders denote our additional/modified components used by the ML pipeline. In (a) the learned model runs on a GPU server that communicates through a TCP socket to an Intel NUC running the flight software. (b) for each path planning iteration, the rover captures a pair of stereo images to generate a height map. The left image is used to generate the traversability estimates, which are projected onto the hazard map using a CAHVORE camera model. (c) a planefit determines the roughness and tilt of cells that are combined with traversability estimates through an averaging into thresholding scheme. The final cost map in (d) is used by the original unmodified search based planner.}
    \label{fig:software}
    \vspace{-1.5em}
\end{figure*}

\textbf{Visual-to-proprio}. The $\mathcal{L}_{v \to p}$ is formed by applying a softmax with the goal of having all $N$ pixels classify which proprioceptive embedding $z_s$ they belong to.
$$
\mathcal{L}_{v \to p} = -\frac{1}{N} \sum_{i=1}^{N}
\log \frac{e^{\ell_{i\sigma(i)}}}{\sum_{s=1}^{S} e^{\ell_{is}}} .
$$
Where $\ell_{is} = \gamma \, f_i \cdot z_s$ represents the similarity between the visual feature $f_i$ and the proprioceptive embedding $z_s$

\textbf{Proprio-to-visual.} Every proprioceptive key runs a softmax over all $N$ pixel queries. A key does not have one correct answer, every pixel of its own segment, is a positive, up to 32 of them. We therefore use multi-positive InfoNCE with SupCon averaging from ~\cite{supinfonce}: for each key, the per-positive log-losses are averaged over its own segment's pixels, and keys whose segment contributed no pixel queries are skipped:
\begin{equation}
\mathcal{L}_{p \to v} = \frac{-1}{|\mathcal{K}^{+}|}
\sum_{s \in \mathcal{K}^{+}} \frac{1}{|\mathcal{P}_s|}
\sum_{i \in \mathcal{P}_s}
\log \frac{e^{\ell_{is}}}{\sum_{j=1}^{N} e^{\ell_{js}}},
\end{equation}
with $\mathcal{K}^{+} = \{ s : \mathcal{P}_s \neq \varnothing \}$. The cross-modal, $\mathcal{L}_{\text{MM}}$, term averages the two directions, and the full training objective adds it to the pixel loss at:
\begin{align}
\mathcal L_{\mathrm{MM}}
 &=\tfrac12(\mathcal L_{v\to p}+\mathcal L_{p\to v}),\\
\mathcal L_{\mathrm{train}}
 &=\mathcal L_{\mathrm{pix}}
   +\lambda_{\mathrm{MM}}\mathcal L_{\mathrm{MM}},
   \qquad \lambda_{\mathrm{MM}}=1.
\label{eq:total_loss}
\end{align}


\subsection{Traversability Prediction}
Since our dataset contains a skewed distribution of slip readings, we propose to learn a quantile head on on top of the transformer as opposed to regression. The quantile head is represented as a two-layer MLP that maps individual pixel features to quantile values: $q_{10}, \ q_{50} \ , q_{90}$. The MLP is trained to minimize the summed pinball loss of unseen (from the perspective of the transformer) sols.
During inference, the final traversability is determined by taking the worst case between the median slip value and the predictive uncertainty (distribution width) described as:
\begin{equation}
    traversability = \max(q_{90} - q_{10},\, q_{50})
\end{equation}
where $q_{50}$ represents the median predicted slip and $q_{90} - q_{10}$ measures the width of the distribution. 
The median penalizes terrain expected to produce high slip, whereas the distribution width penalizes terrain for which the model cannot make a confident prediction. Taking the maximum ensures that terrain is assigned a high traversal cost when either its predicted slip or its uncertainty is high. Consequently, the formulation captures both known high-slip features, such as sand ripples, and uncertain or sparsely observed obstacles, such as rocks and boulders. 
\subsection{Hardware Integration}
We ran experiments on a testbed designed to have the same mobility base as the Perseverance rover, as well as cameras with similar properties. We run a modification of the Mars 2020 Flight Software, \textit{Surface System Development Environment} (SSDev), which provides stubbed software interfaces for hardware interactions.
SSDev was modified with our ML-based path planner on an Intel NUC mounted on top of the rover. Because of compute limitations on the NUC, the model inference was done remotely on a GPU which communicated via a TCP socket to the flight software. The NUC also interacted with the internal rover computer for motor control and sensor data capture.

\subsubsection{Flight Software Integration}
Our approach hooks directly into the existing Mars 2020 Flight Software (FSW), augmenting the existing ENav capabilities, as shown in Figure \ref{fig:software}. We modify ENav to add learned hazard information to the onboard planner. Traditionally, the onboard navigation software maintains two maps: the 2.5D height map and the cost map. The height map is created from disparity images using the public \verb|NAVLIB_process_disparity| interface, which projects a point cloud from the disparity image onto the height map. The 2.5D height map cells are then transferred to a lower-resolution (but much larger) cost map in the \verb|NAVLIB_analyze_terrain| interface, where a plane corresponding to each cost map cell is fit over sections of the height map. 
\begin{figure*}[t!]
    \centering
    \includegraphics[width=0.98\linewidth]{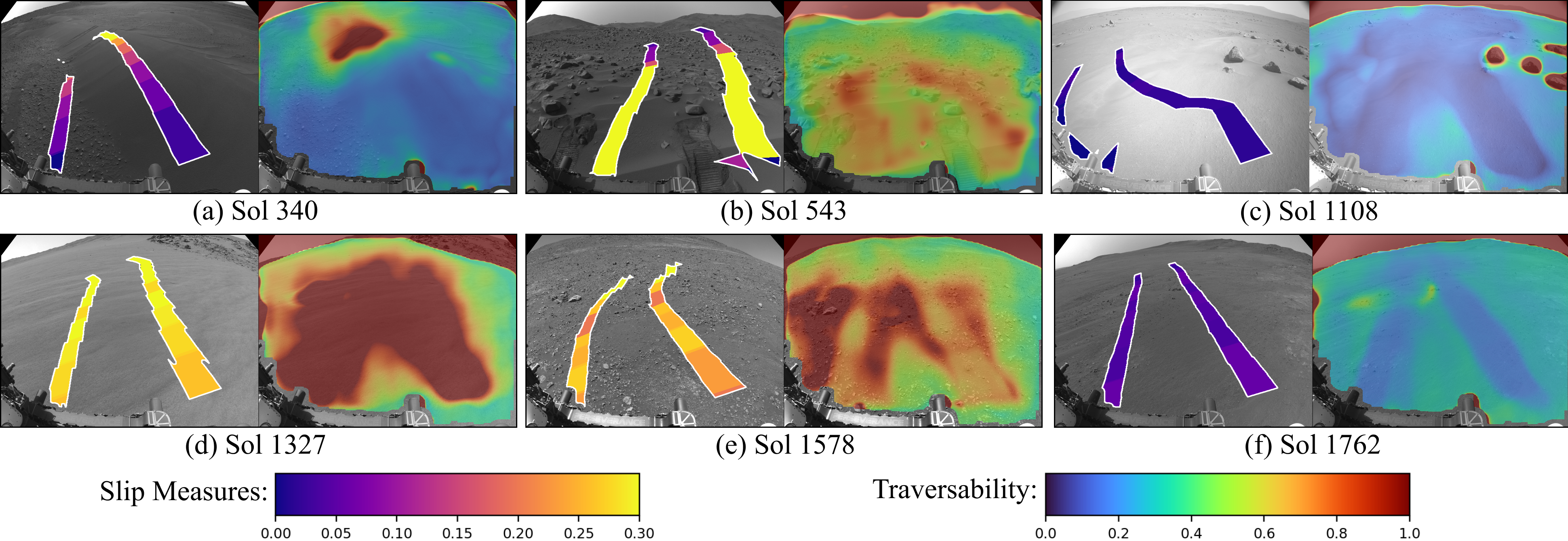}
    \caption{Dataset images and corresponding traversability predictions on unseen held-out test sols. Each pair shows the projected wheel track the rover drove over with its ground-truth slip. The model flags non-geometric sand ripples (a, Sol 340), mixed rock and sand (b, Sol 543), large rocks against safe surroundings (c, Sol 1108), and a sand dune reaching 0.3 slip (d, Sol 1327), while keeping gravel-sand (e, Sol 1578) and flat bedrock (f, Sol 1762) as safe.}
    \label{fig:sim_vis}
    \vspace{-1.0em}
\end{figure*}
Our implementation adds additional functionality to both \verb|NAVLIB| interfaces. Every time the disparity is processed, the FSW connects via a TCP socket to an external ``inference server'' and sends the left NavCam image. The inference server then runs our model over the provided image, and returns another image where each pixel corresponds to a ``hazard'' rating of the corresponding input pixel. 
Once there, the FSW projects this image into a wedge using the calibrated CAHVORE camera model of the original image. Using this wedge, it then encodes each ``hazard'' value into the height map as an attribute alongside the existing ``z'' attribute from the 3D point clouds. At its core, this process adds a third ``hazard'' map to ENav, tracked in parallel to the existing height map.
When \verb|NAVLIB_analyze_terrain| is called, ENav runs the planefit algorithm over a square area of the height map for the cell. Our approach averages the hazard values from each height map cell contained within the area of the cost map cell to generate an ``average hazard'' rating for that cell. In the next step, ENav assigns cost cells a local and global cost based on several features, including tilt, roughness, time to traverse, and more. We have added additional functionality to include the average hazard ratings in these calculations. Cells are also marked as untraversable if they exceed maximum safety values. Additionally, we have created the new \verb|MAX_GLOBAL_HAZARD| fault which occurs when the rover attempts to traverse a cost cell that exceeds the maximum allowed average hazard rating. Later, traversable cell costs are summed into path costs using ENav's path ranking algorithm and are then run through the ACE clearance evaluation algorithm. The added hazard cost directly influences planner path selection, and provides an extra safeguard against visual hazards. 

\begin{table}[t]
\caption{Ablations over the held out test sols. Mean $\pm$ std over 5 seeds.}
\label{tab:ablations}
\centering
\footnotesize
\setlength{\tabcolsep}{3pt}
\resizebox{\columnwidth}{!}{%
\begin{tabular}{@{}lccc@{}}
\toprule
Variant & Hazard AUROC $\uparrow$ & Hazard AP $\uparrow$ & Quantile $\rho$ $\uparrow$ \\
\midrule
\textbf{Full Method} & \textbf{0.874$\pm$ 0.003} & \textbf{0.878 $\pm$ 0.004} & \textbf{0.721 $\pm$ 0.007} \\
$\lambda_{\text{MM}} = 0$ & 0.723 $\pm$ 0.022 & 0.745 $\pm$ 0.016 & 0.316 $\pm$ 0.051 \\
No Rank-N-Contrast \cite{zha2023rank} & 0.778 $\pm$ 0.009 & 0.765 $\pm$ 0.017 & 0.703 $\pm$ 0.007 \\
\bottomrule
\end{tabular}}
\vspace{-1.5em}
\end{table}
\subsubsection{FSW Parameters and Configuration}
Our ML-based planning approach can be tuned to specific scenarios with several parameters. In general, these mirror existing ENav parameters for similar features. Unseen cost map cells are initialized with the \verb|unknown_avg_hazard| parameter, which represents the average hazard rating of an unknown cell. Additionally, the \verb|max_global_avg_hazard| parameter controls the maximum safe hazard threshold the rover can traverse before triggering a \verb|MAX_GLOBAL_HAZARD| fault. Finally, hazard values can only begin influencing local cost when they exceed \verb|avg_hazard_cost_threshold|, at which point their values (minus the threshold) are multiplied by \verb|avg_hazard_cost_weight| to calculate the total added hazard cost.

           
\section{Results}

\begin{table}[t]
\caption{Comparison to baselines on the held out test sols.
Metrics are reported with a mean $\pm$ std over 5 seeds.}
\label{tab:baselines}
\centering
\footnotesize
\setlength{\tabcolsep}{3pt}
\resizebox{\columnwidth}{!}{%
\begin{tabular}{@{}lccccc@{}}
\toprule
Method & AUROC $\uparrow$ & AP $\uparrow$ & Prec $\uparrow$ & Recall $\uparrow$ & F1 $\uparrow$ \\
\midrule
V-STRONG \cite{vstrong} & 0.645 $\pm$ 0.016 & 0.670 $\pm$ 0.014 & \textbf{0.958 $\pm$ 0.010} & 0.118 $\pm$ 0.011 & 0.210 $\pm$ 0.017 \\
STERLING \cite{karnan2023sterling} & 0.636 $\pm$ 0.010 & 0.612 $\pm$ 0.012 & 0.751 $\pm$ 0.088 & 0.080 $\pm$ 0.054 & 0.137 $\pm$ 0.089 \\
Castro et al. \cite{Castro_2023_HowDoesItFeel} & 0.816 $\pm$ 0.013 & 0.802 $\pm$ 0.019 & 0.794 $\pm$ 0.017 & 0.486 $\pm$ 0.034 & 0.602 $\pm$ 0.030 \\
\midrule
\textbf{Ours} & \textbf{0.874 $\pm$ 0.003} & \textbf{0.878 $\pm$ 0.004} & 0.835 $\pm$ 0.011 & \textbf{0.695 $\pm$ 0.020} & \textbf{0.758 $\pm$ 0.008} \\
\bottomrule
\end{tabular}}
\vspace{-1.5em}
\end{table}

\label{sec:results}
\begin{figure*}[!tbp]
    \centering
    \includegraphics[width=0.98\linewidth]{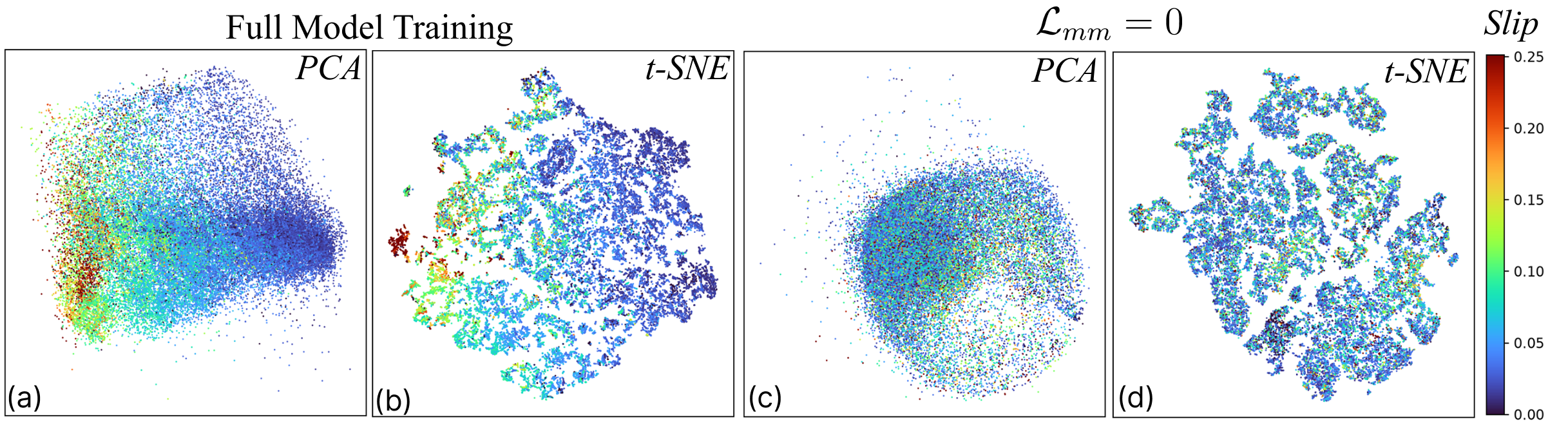}
    \caption{Latent visualization of the full framework via (a) PCA and (b) t-SNE; (c) and (d) show the same without multimodal alignment $\mathcal{L}_{mm} = 0$.Colors denote ground-truth slip from the driven pixels in our dataset.}
    \label{fig:latent}
    \vspace{-0.0em}
\end{figure*}
\begin{figure*}[h!]
    \centering
    \includegraphics[width=0.98\linewidth]{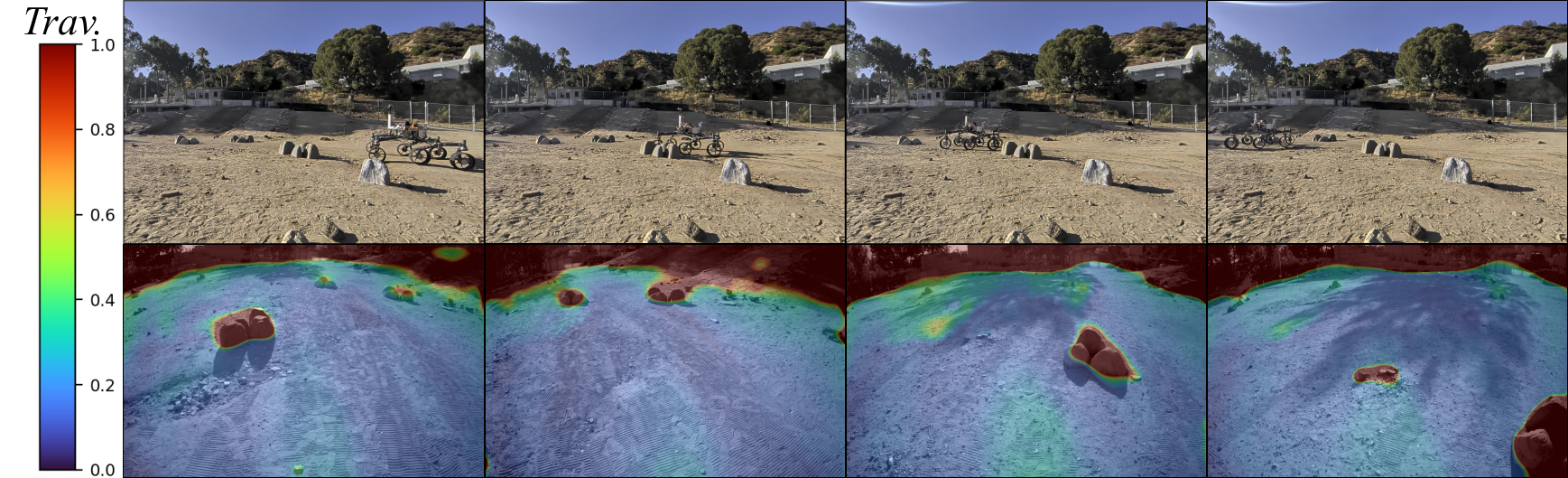}
    \caption{Navigation around several boulders using a rover test bed. The top row shows the rover successfully planing around the obstacles. The bottom row visualizes the traversability estimation from the learned model.}
    \label{fig:hardware}
    \vspace{-1.5em}
\end{figure*}
Our experiments are conducted on 35 held-out test sols of driving, where we separate safe and unsafe pixels by a threshold over the slip measures $90^{\text{th}}$ percentile along with depth protruded rocks. This lets us measure binary metrics such as area under the receiver operating characteristic curve (AUROC) and average precision (AP). We also report the Spearman rank correlation ($\rho$) between the learned median $q_{50}$ and the slip labels from the driven footprints. Visualizations under the held out test sols are shown in Figure \ref{fig:sim_vis}.            
\subsection{Ablation Studies}
Table \ref{tab:ablations} evaluates the contributions of the two training objectives while retraining the same inference-time worst case traversability readout. The full method achieves the highest mean performance across the three reported metrics. Removing the cross-modal alignment ($\lambda_{\text{MM}}=0$) produces the largest degradation in overall performance. The effect of this ablation is especially pronounced under the Spearman rank $\rho$. This indicates that slip is not solely predictable from visual features and requires other information such as the vehicles tilt and suspension angles. Overall, the ablations support complementary roles for the final objective and show that their combination provides the strongest performance under our metrics. 
\subsection{Baseline Comparisons}
Table \ref{tab:baselines} compares our method with V-Strong \cite{vstrong}, STERLING \cite{karnan2023sterling}, and How Does it Feel \cite{Castro_2023_HowDoesItFeel} (denoted by Castro in Table \ref{tab:baselines}). We show that on the M2020 dataset, the proposed method scores the highest on AUROC, AP, Recall, and maximum F1. V-Strong scores higher on Precision, but substantially lower on other metrics due to the methods approach of using the vehicle tracks as a much harsher contrastive loss (i.e. marking non driven pixels as untraversable). Our method looks to rank pixels \textit{within} the tracks in addition to creating contrast on non driven pixels. 
STERLING obtains an AUROC of 0.636, an AP of 0.612, and F1 of 0.137. STERLING's native VICReg-based formulation is intended to learn transferable terrain representations by associating visual and proprioceptive observations. In our adaptation, the descriptors contain information about vehicle response, but the non-contrastive association objective does not explicitly require high-slip terrain to be separated from low-slip terrain or arranged along a monotonic slip direction (shown in Figure \ref{fig:latent}). Different slip values can therefore remain close in the learned feature space as long as the paired visual and proprioceptive representations agree. This limitation is amplified by our sparse, segment-level supervision, for which a single chassis-level response is assigned to multiple pixels along the projected wheel tracks. 
The Castro-style baseline is better aligned with the task and is consequently the strongest comparison, attaining an AUROC of 0.816 and an AP of 0.802. Slip is, in principle, more aligned to Castro's scalar-regression formulation because it provides a continuous target analogous to the vibration cost used in the original method. However, direct regression is not well matched to the skewed distribution of our labels. Most observations have low slip, whereas the high-slip tail is comparatively sparse. A pointwise regression loss is therefore dominated by the abundant low-slip samples, and reducing absolute prediction error does not necessarily preserve the relative ordering of the high-slip examples that determine AUROC and AP. 
\subsection{Latent Visualization}

Figure ~\ref{fig:latent} projects the 64-dimensional features of held-out wheel-track pixels into two dimensions, colored by ground-truth slip. Panels~(a),(c) use PCA, which preserves dominant global directions of variation; panels~(b),(d) use t-SNE, which emphasizes local neighborhood structure. Both use the same samples.

The PCA projection exhibits a continuous color gradient, with high-slip samples concentrated on the left and low-slip on the right. The representation therefore does not separate the data into "safe" and "hazardous" clusters; instead, one of its dominant directions varies continuously with slip magnitude. The t-SNE projection shows complementary local structure: neighboring points share similar color while preserving the gradient. High-slip samples form multiple local groups rather than one isolated cluster, suggesting that elevated slip corresponds to more than one visual terrain appearance, while the coherence of the low-slip region on the right indicates a consistent representation of benign driven terrain.
\subsection{Hardware Experiments}
We show that the proposed method is able to transfer towards a visually out of distribution environment on an Earth analog testing site. We performed testing primarily on flat ground, with a variety of large rocks serving as obstacles. Figure \ref{fig:hardware} shows still shots as the rover navigates around various boulders. A traversability map is generated by the model and projected onto a wedge. Our modified planner accepts the traversability scores and fuses them with the original geometric based cost map to search a path over.
\section{Conclusion}
We present a dataset and learning framework for dense, vision-only traversability prediction from Perseverance data. By projecting the rover's wheel tracks into pre-traversal NavCam imagery, the method contrastively ranks visual features against proprioception and suspension response without requiring manual annotation. We train a dense decoder over features generated from a DINOv2 encoder using complementary objectives: a Rank-N-Contrast loss structures features within the driven region, while bidirectional cross-modal alignment relates those features to the rover's mechanical response. Future work could support online adaptation to new terrains or scale the method towards larger backbones such as a VLM. 
\ifanon\else
\section{Acknowledgements}
This research was carried out at the Jet Propulsion Laboratory (JPL), California Institute of Technology, under a contract with the National Aeronautics and Space Administration (80NM0018D0004). This work was supported by JPL’s Strategic Research and Technology Development (SRTD) program. The High Performance Computing resources used in this investigation were provided by funding from the JPL Enterprise IT Division. The authors would like to thank Tyler Del Sesto and Richard Rieber for their insightful discussions on the hardware platform and ENav algorithm, along with JPL's Rover Operations Center (ROC) for their support. 
\fi

\bibliographystyle{IEEEtran}
\bibliography{citations.bib}
\end{document}